\documentclass[letterpaper, 10 pt, conference]{ieeeconf}  

\IEEEoverridecommandlockouts                              

\usepackage{graphics} 

\usepackage{graphicx}
\usepackage{caption}
\usepackage{subcaption}
\usepackage{color,soul}
\usepackage{amsmath}
\usepackage{siunitx}
\usepackage{epstopdf}

\usepackage{epsfig} 
\usepackage{mathptmx} 
\usepackage{times} 
\usepackage{amsmath} 
\usepackage{amssymb}  

\title{\LARGE \bf
Design, Fabrication and Control of an Hydraulic Elastomer Actuator
}

\author{Mahdi Momeni Kelageri, Mikko Heikkil{\"a}, Minna Poikelisp{\"a}{\"a}, Reza Ghabcheloo, Matti Linjama, Jyrki Vuorinen$^{1}$
\thanks{*This work was supported by Academy of Finland}
\thanks{$^{1}$Authors are with Tampere University of Technology, Tampere, Finland
        {\tt\small \{mahdi.momenikelageri, mikko.heikkila, reza.ghabcheloo, matti.linjama, minna.poikelispaa, jyrki.vuorinen\}@tut.fi}}%
}

\begin{document}

\maketitle
\thispagestyle{empty}
\pagestyle{empty}

\begin{abstract}

This paper presents design, fabrication and control of a compliant 2D manipulator, a so called soft actuator. Our focus is on fiber-reinforced elastomer actuators driven by a constant pressure hydraulic supply and modulated on/off valves. For a given diameters, we study the effect of four different elastomer materials and that of number of reinforcement fiber turns on forces generated by the actuator and maximum bending angles. For the rest of the study, we use polydimethylosiloxane (PDMS) with $240$ fiber turns per $170~mm$ length of actuator which withstand highest pressures and forces in our experiments. For the rest of the paper, we introduce two control methodologies. Firstly, we show that is possible to reasonably accurately control the pressure inside tube without measuring the pressure incorporating a simple linear tube model. This can be used, for example, in an inner-outer loop configuration with a PI position control to achieve high performance without the need for pressure measurement. Secondly, we experimentally show that a switching position control exhibits very good steady state accuracy and acceptable transient. Actuator tip position is measured using an external vision system. Our experiments included performance analysis of our soft manipulator while freely moving as well as when carrying a load. 

\end{abstract}

\section{INTRODUCTION}

Engineers have long been inspired by biology in order to make ever-more capable machines. Of noticeable features exploited from biological systems is softness and body compliance which tend to seek simplicity in interaction of such systems with their environment. Several of the lessons learned from studying biological systems are now culminating in the definition of a new class of machines that is referred to as soft robots \cite{Daniela2015_Nature}. While traditional rigid robots are conventionally used in manufacturing tasks where performing a single task efficiently is desired, soft robots can serve better in human-centric operations where safety and adaptability to uncertainties are fundamental requirements \cite{Daniela2015_IJRR}.

Soft robots are made of continuously deformable materials with, theoretically, infinite number of configurations. This means that the robot tip can attain every point in $3$D workspace with an infinite number of configurations. In addition to that, soft robots are capable of carrying soft and fragile payloads without causing any damages \cite{Trivedi2008, Chen2006}. These fundamental properties allow soft robots to: $1)$ adaptively grasp and manipulate unknown objects with different shapes and sizes \cite{McMahan2006}, and $2)$ squeeze through confined spaces \cite{Shepherd2011}.

In traditional rigid link robots, the position of the end-effector is calculated by measuring the position of each joint through several high resolution encoder. By using forward kinematics the orientation and position of the robot tip can be determined with high accuracy. On the other hand, inverse kinematics can be used to determine the joint variables corresponding to the desired tip pose. Soft robots, however, interact with the environment quite differently. They have a continuum structure rather than well defined joints and links, and they articulate by means of material compliance. As a matter of fact, shape estimation \cite{Trivedi2014} is an important challenge in soft robots. One common approach to shape estimation in continuum robots is measuring strain along the manipulator axis \cite{Leleu2001, Lyshevski2003} and applying dynamical models of the manipulator in order to predict the curvature of the manipulator based on measured strain. Another method for shape estimation is using fiber optic sensors along the body. This method, however, suffers from propagation losses when they are bent \cite{Miller2004}. In this paper, we use vision to estimate the tip position and curvature of the actuator similar to \cite{Hannan2005}.

Another important challenge in soft robotics is the development of controllable soft bodies using materials that integrate sensors, actuators and computation, and that together enable the body to deliver the desired behavior \cite{Daniela2015_Nature}. It should also be noted that soft robots are classified as under-actuated systems because, unlike their traditional rigid link robots counterparts, there is not an actuator for every degree of freedom \cite{Trivedi2008}. Furthermore, gravity and/or payload manipulation cause continuous deformation in soft bodied robots that may not be observable and/or controllable from the limited sensors or actuators.

\begin{figure}
    \centering
    \begin{subfigure}[t]{0.23\textwidth}
        \includegraphics[width=\textwidth]{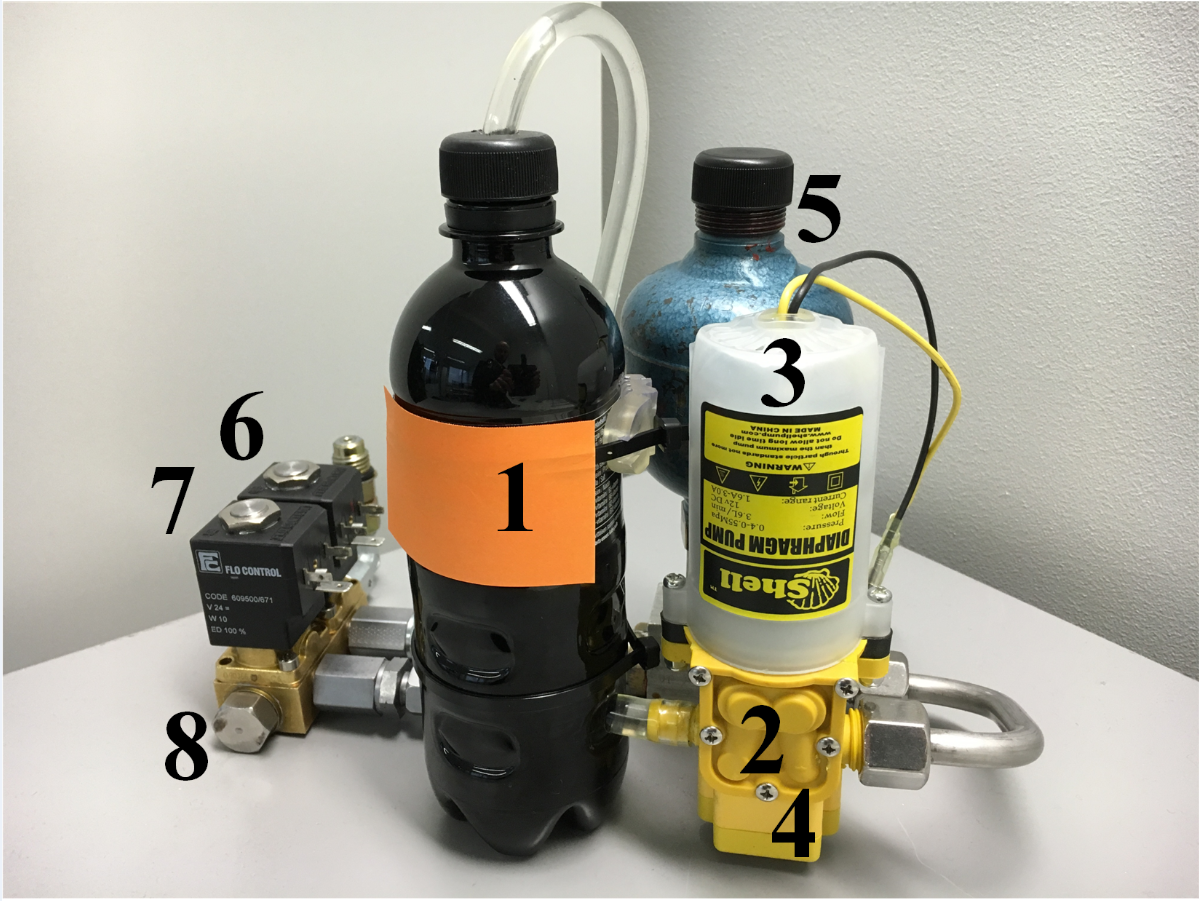}
        \caption{test system}
        \label{fig:DHTestSystem}
    \end{subfigure}
    ~ 
    \begin{subfigure}[t]{0.23\textwidth}
        \includegraphics[width=\textwidth]{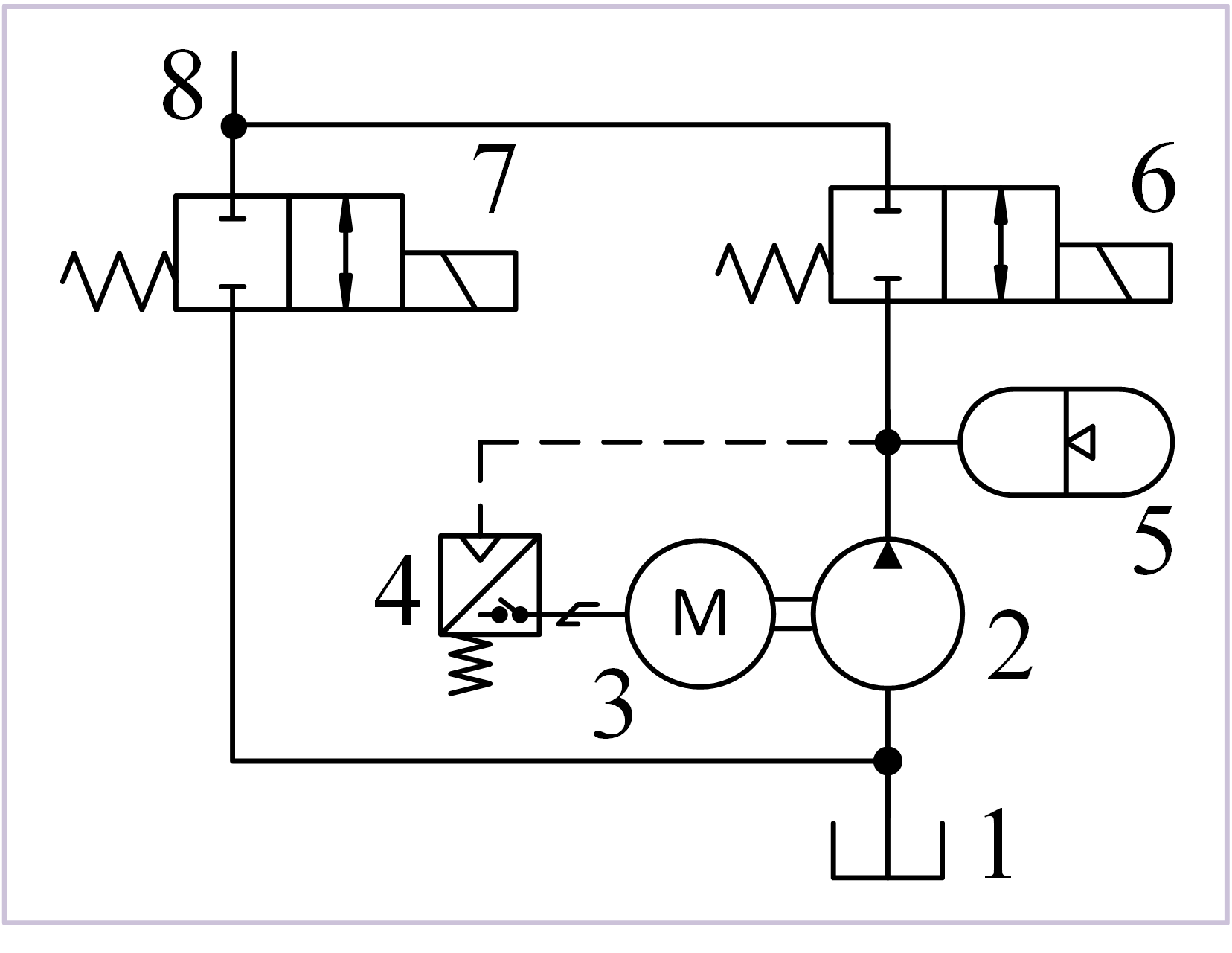}
        \caption{schematic diagram}
        \label{fig:DHSchematicDiagram}
    \end{subfigure}
    ~ 
    \caption{Digital hydraulic drive system for controlling a soft actuator}
    \label{fig:DigitalHydraulic}
\end{figure}

In general, the design of soft robots are not very straight forward. Some soft robots use tendons for pulling rigid fixtures embedded on their body as their actuation scheme \cite{Gravagne2002}. In \cite{Wang2013} position of a soft rubber arm was controlled using cables without rigid plates. Distributed pneumatic muscle actuators (PMAs) are another common design of position controlled soft manipulators \cite{McMahan2006}. Many different types of soft actuators working under pressure have been developed including e.g., McKibben actuators \cite{Pritts2004}, PneuNets (Pneumatic Networks), and fiber-reinforced actuators \cite{Daniela2015_Nature, DeGreef2009, McEvoy2015}. Operating principle of fiber-reinforced actuators is based on anisotropic structure and expansion in the direction of the lowest modulus enabling wide range of motions \cite{Polygerinos2015, Galloway2013}. Fiber-reinforced actuators are studied by many researchers. According to these studies, there are many variables affecting actuation results such as shape, length, inner radius, outer radius, wall thickness, number of fiber turns and fiber angle \cite{Polygerinos2015}\nocite{Galloway2013}-\cite{Bishop2012}. However, the effect of different elastomer types is rarely studied. In most cases soft actuators are fabricated from silicone \cite{Polygerinos2015}\nocite{Galloway2013}-\nocite{Bishop2012}\cite{Gorissen2014} due to its softness and ease of use. In addition, use of latex has been reported in \cite{Bishop2012}. There are many other potential elastomers that has not been yet studied which are able to withstand higher pressures and produce higher force outputs. In this paper, the effect of different elastomer types on actuation performance of fiber-reinforced actuators is studied. Furthermore, the digital hydraulic drive system and a model-based controller with quite fast response is presented to control the tube pressure without the need for measuring the pressure. They are all experimentally validated by controlling the position/pressure of the actuator while carrying a payload and moving freely. We will also experimentally validate a switching control strategy.

This paper is organized as follows: An introduction to soft robots was given in Section I. Section II details the entire system. System modelling and control is presented in Section III while section IV presents the experimental validations. Eventually, the research is discussed and concluded in section V.

\section{System Description}

The hydraulic elastomer actuator system studied in this paper is composed of: $A)$ a soft and compliant one-directional elastomer tube, $B)$ digital hydraulic drive system, $C)$ vision system for curvature estimation and tip localization, Fig.~\ref{fig:DigitalHydraulic} and Fig.~\ref{fig:ExpSetup}, and $D)$ control

\subsection{Elastomer Tube}

In this work, the manipulator's workspace is constrained to X-Y plane while it is only capable of one-directional bending. Bending principle is based on anisotropic structure and expansion in the direction of the lowest modulus by pressurizing or depressurizing internal fluid which induces stress in elastomer. 

\begin{figure}
    \centering
        \includegraphics[trim={0 0 0 0},clip, width=8.0 cm]{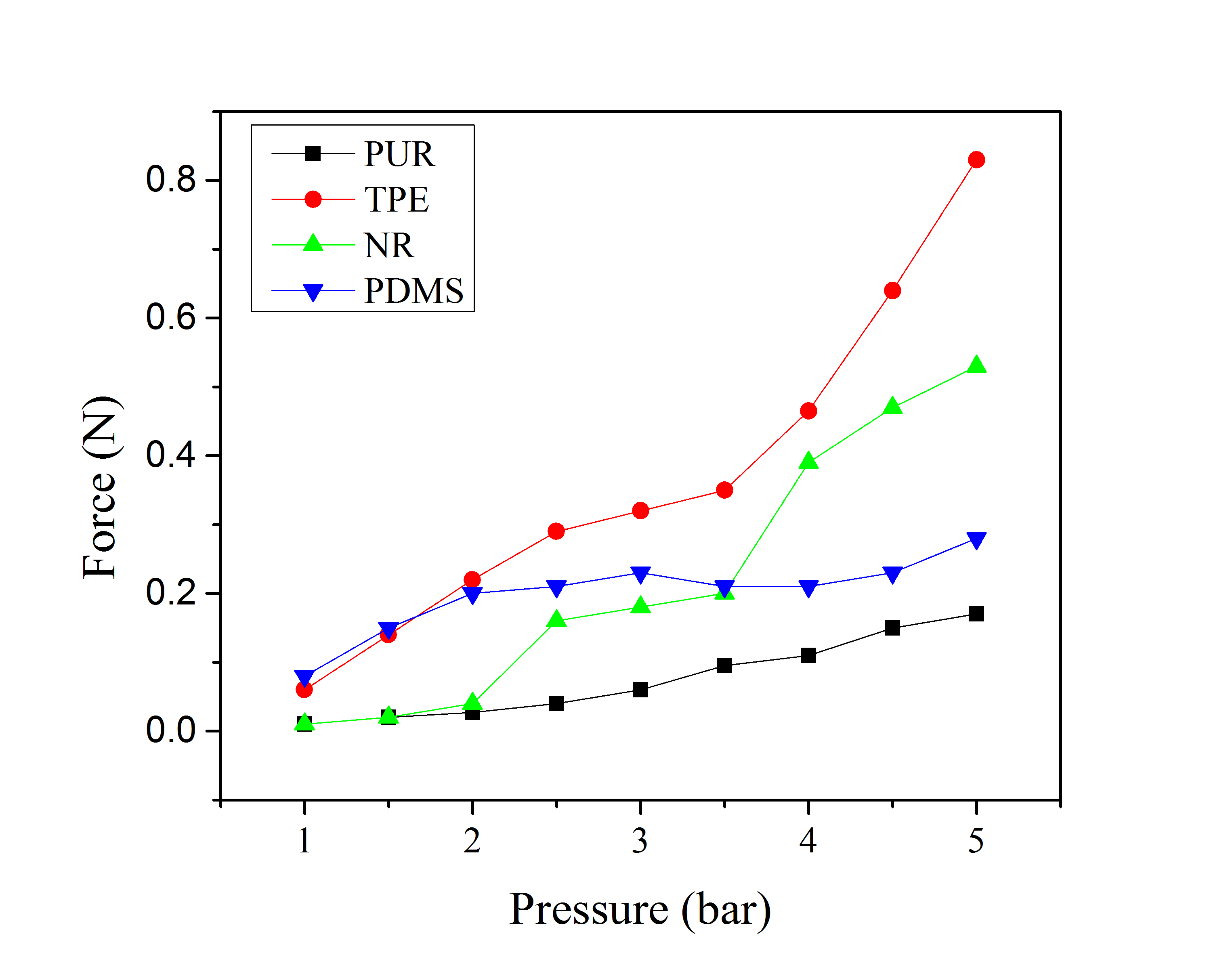}
        \caption{Force output of the actuators fabricated from different elastomers}
        \label{fig:Force-Pressure}
\end{figure}

\begin{figure}
        \includegraphics[trim={0 0 0 0},clip, width=8.0 cm]{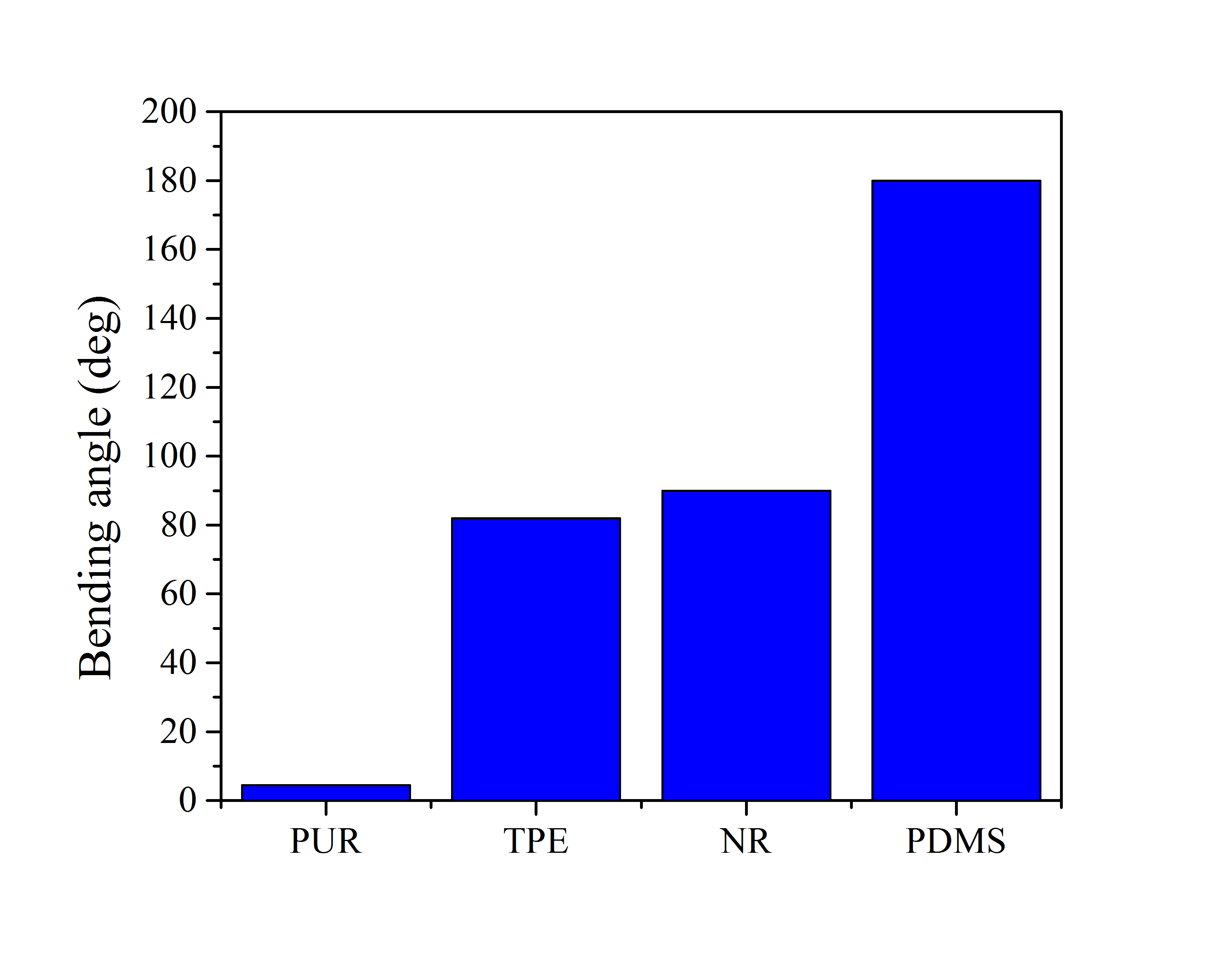}
        \caption{Bending angle at 5$~KPa$}
        \label{fig:Bending-Elastomer}
\end{figure}

\begin{figure}
    \centering
        \includegraphics[trim={0 0 0 0},clip, width=8.0 cm]{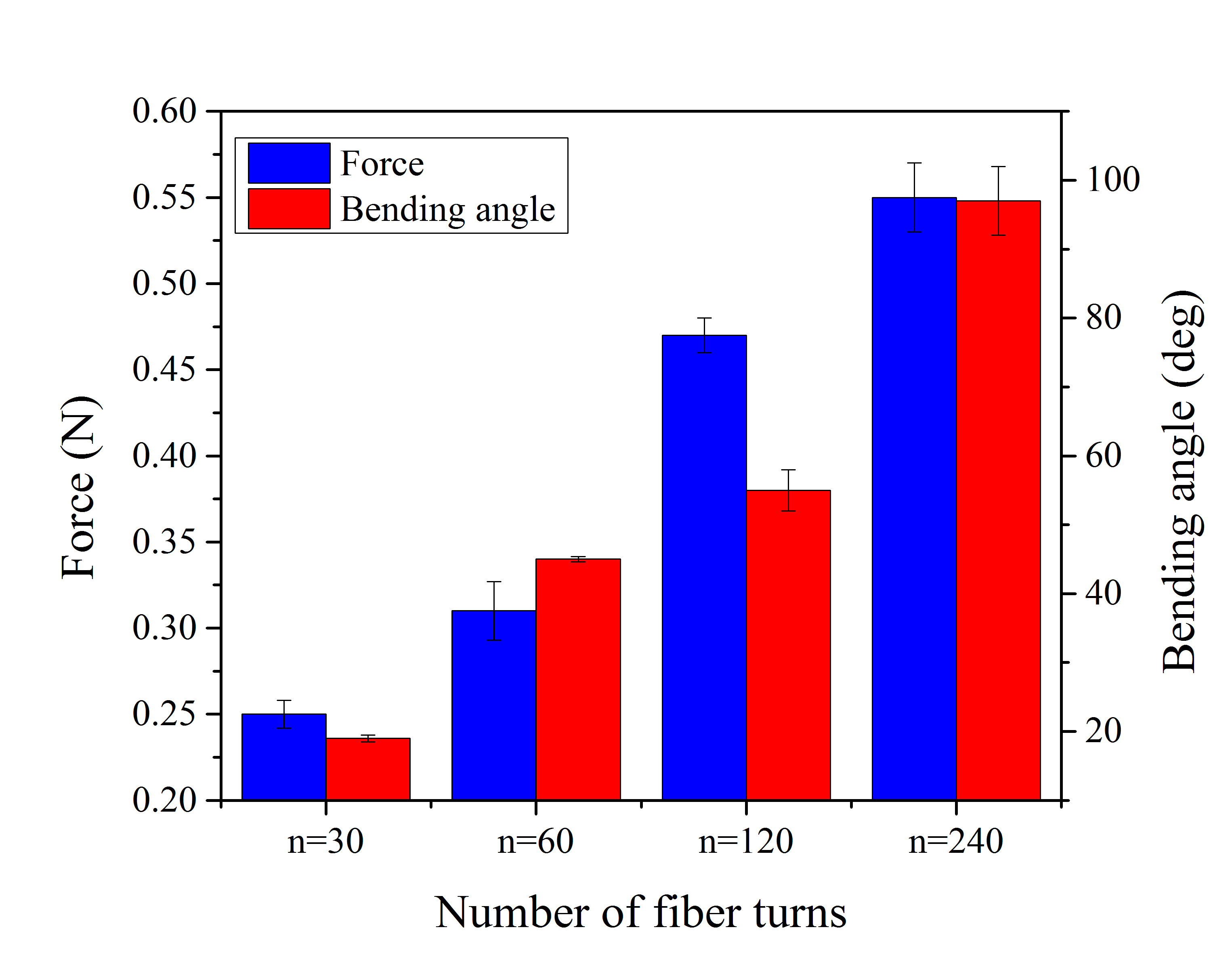}
        \caption{The effect of number of turns on actuation performance of PDMS at 5$~KPa$}
        \label{fig:ActuationPerformance-Turns}
\end{figure}

To study the effect of different elastomer types, four different elastomers have been prepared (please refer to APPENDIX) and their performance have been analyzed. Fig. \ref{fig:Force-Pressure} and \ref{fig:Bending-Elastomer} show different actuators performance. PDMS shows higher bending angle, Fig.~\ref{fig:ActuationPerformance-Turns} and thus, satisfies our requirements better. For control part of this research, number of turns of fibers was set to $240$ as it gave the highest force output as well as bending angle.

\subsection{Digital Hydraulic Drive}

A digital hydraulic drive system is used to control the soft robotic actuator. The test system is shown in Fig.~\ref{fig:DHTestSystem} while the corresponding hydraulic diagram is depicted in Fig.~\ref{fig:DHSchematicDiagram}: The size of the tank 1 is $0.5~l$ and it is connected to the diaphragm pump 2. The pump is then connected to a $12~V DC$ motor 3 having a maximum power of $36~W$. The maximum flow rate volume of the hydraulic power unit 2 is $3.6~l/min$ whereas the maximum system pressure is limited to about $600~kPa$ by the pressure switch 4. The hydro-pneumatic accumulator 5 is attached to the supply line to store the hydraulic energy. The fluid volume in the actuator port 8 can be increased by opening the high-pressure valve 6. On the other hand, opening the low-pressure valve 7 decreases the actuator fluid volume as the flow direction is towards the tank. The orifice diameter of these on/off valves is around $0.7~mm$. Water is used as the hydraulic medium in the drive system.

\subsection{Vision System}

A single Microsoft LifeCam camera has been used in order to find out the curvature of the manipulator and localize the tip. The resolution of the camera is set to $800\times600$, while MATLAB Image Processing and Computer Vision Toolboxes have been used for tip localization and curvature estimation software. The camera is also calibrated at the initialization time before the experiment using the partial chess board attached behind the manipulator.

\subsection{Contol Hardware}

A dSpace MicroAutobox controls the valves through digital and analog I/O and runs pressure control algorithms, a windows PC runs image processing functions. They are communicating to each other through CAN bus. The communication frequency is $5~ms$ and the overall delay is $4~ms$, $2~ms$ in sending the commands to dSpace and $2~ms$ in reading the data from dSpace.

\section{Modelling and Control}

In this chapter, a simulation model is created for the studied system and the basic control principle for a soft manipulator is described when the digital hydraulic drive system is used. In addition, a model-based approach for the open-loop pressure control is proposed based on the simulation model. A simulation shows the feasibility of this approach in ideal case. A method for the closed-loop position control is also presented.

\subsection{Simulation Study}

\begin{figure}[t]
\includegraphics{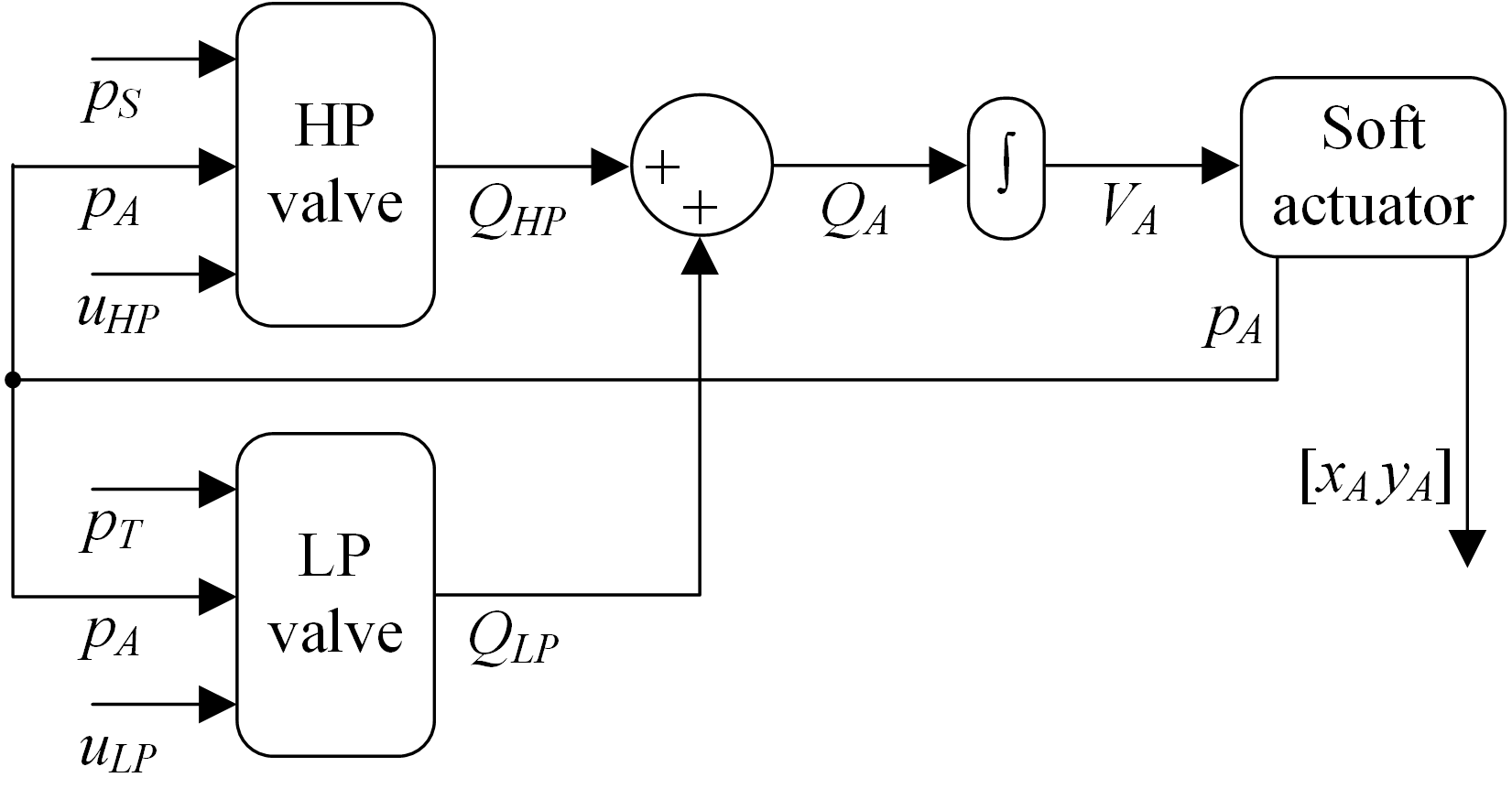}
\caption{A block diagram of the studied system.}
\label{fig:sim_model}
\end{figure}

MATLAB/Simulink is used to create a simple model for the studied system. Fig.~\ref{fig:sim_model} shows a block diagram for the model. Orifices of the high pressure and low pressure valves are modeled based on the turbulent volume flow. In addition the flow model mimics the laminar flow when the pressure difference over the orifice is small. Thus, infinite derivative can be avoided when the pressure difference equals zero \cite{ellman_1999}. The equations can be written as follows:

\begin{equation}
\label{eq:flow_orif}
Q = 
\begin{cases}
\displaystyle op \cdot K_{v} \cdot sgn \left( p_{1} - p_{2} \right) \cdot \sqrt{\left| p_{1} - p_{2} \right|} & : \left| p_{1} - p_{2} \right| > p_{tr} \\[0.15cm]
\displaystyle op \cdot \frac{K_{v} \cdot \left( p_{1} - p_{2} \right)}{2 \cdot \sqrt{p_{tr}}} \cdot \left( 3 - \frac{ \left| p_{1} - p_{2} \right|}{p_{tr}} \right) & : \left| p_{1} - p_{2} \right| \leq p_{tr} 
\end{cases}
\end{equation}

where $K_{v}$ is the orifice flow factor determined from the nominal volume flow and pressure. The transition pressure between the turbulent and laminar flow is determined by $p_{tr}$. For the high pressure valve $p_{1} = p_{S}$ and $p_{2} = p_{A}$. Correspondingly, $p_{1} = p_{T}$ and $p_{2} = p_{A}$ for the low-pressure valve. The valve opening $op$ can have a value between $0$ and $1$. Opening and closing dynamics of the valve armature are modeled based on three parameters: 1) delay, 2) movement time, and 3) sticking time as detailed in \cite{Schepers_2011}. This model describes realistically the valve operation in practically any condition. The control signals  $u_{HP}$ and $u_{LP}$ can have a value either $0$ or $1$.

The volume flow for the actuator is calculated as a sum of flows: $Q_{A} = Q_{HP} + Q_{LP}$. This volume flow is further integrated over time to achieve the fluid volume inside the actuator. The used actuator model is a linear function: $p_{A} = C_{A} \cdot V_{A}$, describing the relation between the fluid volume and pressure inside the tube ($C_{A}$ is a constant). Thus, dynamic properties of the actuator are left aside as are the non-linearities of a rubber material.

\subsection{Basic Control Principle}

Primarily, the fluid volume inside the soft robotic actuator can be influenced by controlling the on/off valves. The fluid volume can be increased by opening the high-pressure valve and decreased by opening the low pressure valve. Consequently, the fluid volume affects the pressure inside the soft actuator. Furthermore, the pressure determines the shape of the actuator.

\subsection{Model-Based Pressure Control}

In this section we study how we can control the pressure inside the tube using the simulation model, without utilizing the direct pressure feed-back. The controller utilizes a model for the high-pressure and low-pressure valves based on the equation of turbulent flow (\ref{eq:flow_orif}). The anticipated fluid volume passed through the valves is calculated for each time step (sampling period). The assumption is that each the supply pressure, tank pressure, and the pressure inside the tube remain constant during the sampling period. 

The actuator model has the total fluid volume inside the tube as an input, whereas the output is the corresponding pressure ($C_{A} = 3.3 \cdot 10^{11}~Pa/m^{3}$). Thus, the estimated error in the pressure is calculated for every feasible control combination: 1) HP-valve OFF and LP-valve OFF, 2) HP-valve ON and LP-valve OFF, and 3) HP-valve OFF and LP-valve ON. The control combination which minimizes the error between the desired and estimated pressure is then fed to the system. In addition, a tolerance value for the pressure error can be used to avoid excessive valve switchings. Finally, the new values for the actuator pressure and fluid volume are estimated in order to be used during the next time step. 

\begin{figure}[t]
\includegraphics{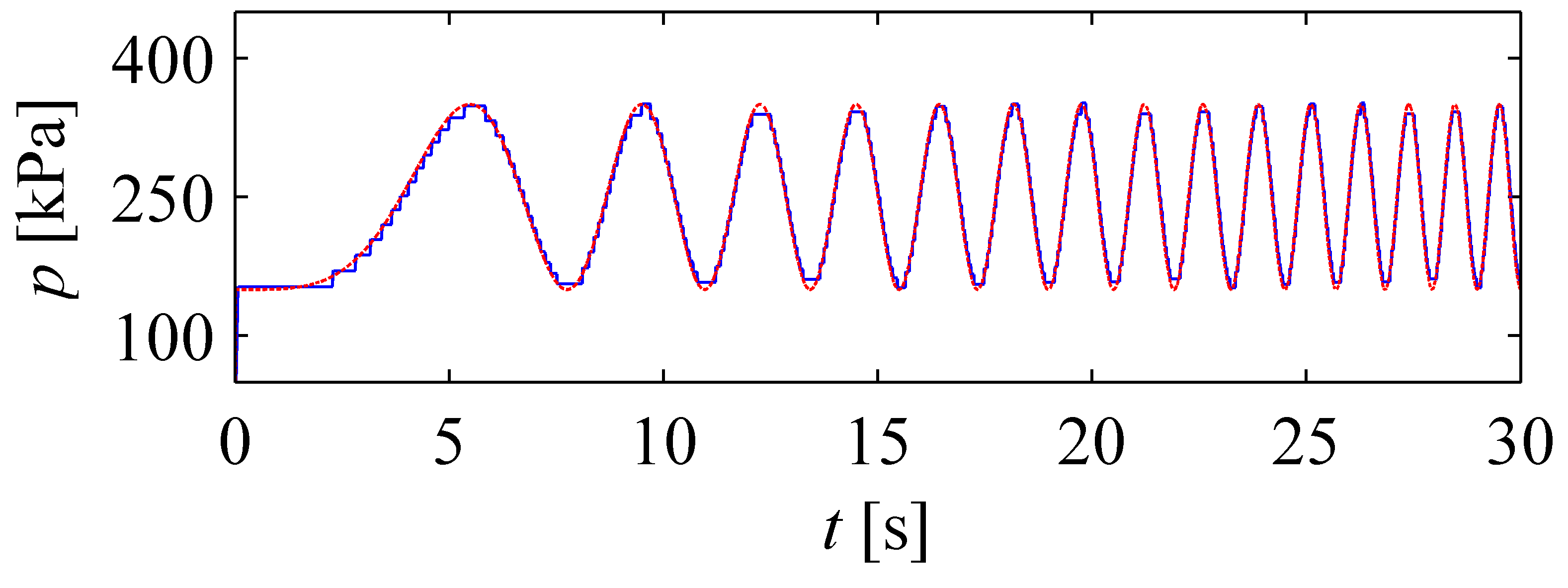}
\caption{Open-loop pressure control response: a simulation using a linear model for the soft robotic actuator.}
\label{fig:mbc_sim}
\end{figure}

Fig.\ref{fig:mbc_sim} shows the response to a sinusoidal reference signal having an increasing frequency from $0$ to $1~Hz$. The minimum value of the pressure reference is $150~KPa$, whereas the maximum is $250~KPa$. The simulation shows that the actuator pressure can be controlled quite accurately when the system parameters are known. However, the control resulution is low due to slow response time of the valves: the shortest acceptable duration for the opening/closing command is $5~ms$. The error tolerance was set to $10~KPa$.

\subsection{Switching Position Control}

The controller is a switch that acts like a signum function with sampling time equal to $100~ms$, where the control input is the position error $e_P$:

\begin{equation}
\label{eq:sgnfun}
sgn = 
\begin{cases}
\displaystyle		1	&	 : threshold > e_p \\[0.15cm]
\displaystyle		0	&	 : -threshold < e_p < threshold\\
\displaystyle		-1	&	 :   e_p < -threshold
\end{cases}
\end{equation}

In other words, the switching controller switches the high pressure valve on if $u = 1$, and low pressure valve on if $u = -1$ and turns both valves off if $u = 0$. During the experiment, we found that the existing valve capacity was too high and made the manipulator tip move too fast to be tracked by the camera. Therefore, we decided to modulate the valve control signals to get only 15-22\% of their capacity.


\section{Experiments}

\begin{figure}[t]
\centering
\includegraphics[trim={0 0 0 0},clip, width=8.0 cm]{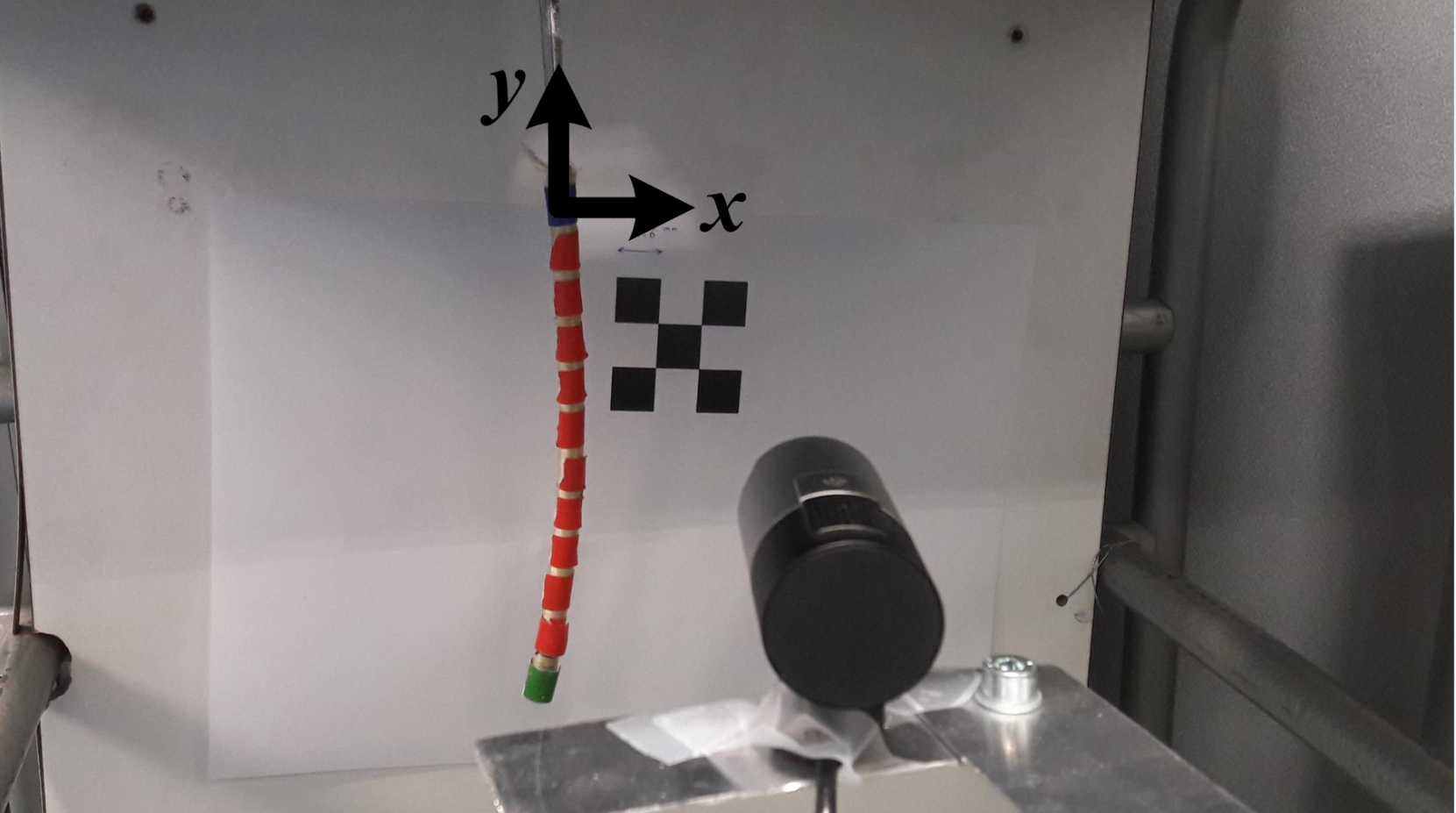}
\caption{Experimental Setup}
\label{fig:ExpSetup}
\end{figure}

The current version of the soft manipulator, Fig.~\ref{fig:ExpSetup}, is capable of moving payloads up to $30~g$. The manipulation system can accurately control the tip position and pressure of the soft robot in real-time and move the tip to any desired position in its workspace. In this section we show the performance of the soft manipulator in both loaded and unloaded scenarios. The main objective of these exeriments is to control the position of the robot tip and find out the relationship between tip position and tube pressure. To this end, a model-based approach for controlling the pressure (without measuring the pressure) has been developed. The experiments show that using this model, it is possible to estimate and control the tube pressure with acceptable accuracy. It will also be experimentally shown that due to the high nonlinearities in the existing hysteresis of the system, there is not any one-to-one correspondence between tip position and tube pressure unless the initial conditions are the same. That is to say it is not possible to accurately control the tip position with only a pressure feedback (without position feedback) unless the initial conditions are the same.


\subsection{Model-Based Pressure Control}

\begin{figure}[t]
\includegraphics{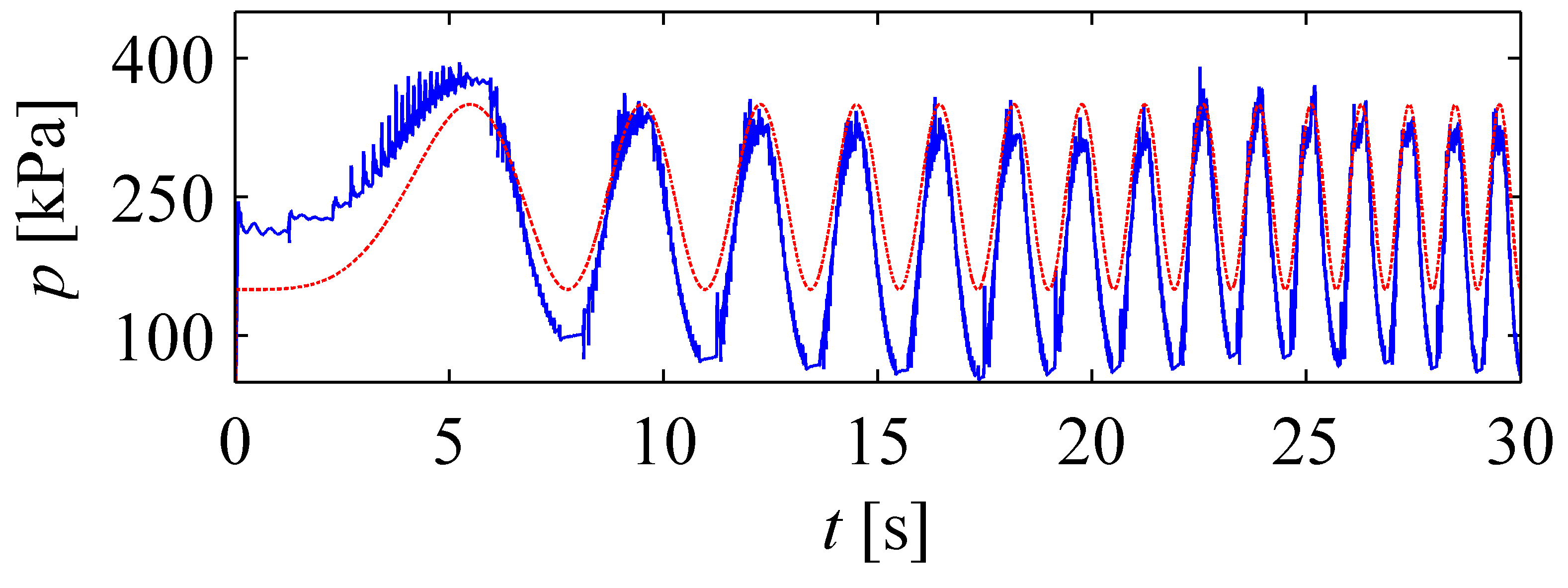}
\caption{Open-loop pressure control response: a measurement using  a linear model for the soft robotic actuator and the presumable valve parameters.}
\label{fig:mbc_meas1}
\end{figure}

\begin{figure}[t]
\includegraphics{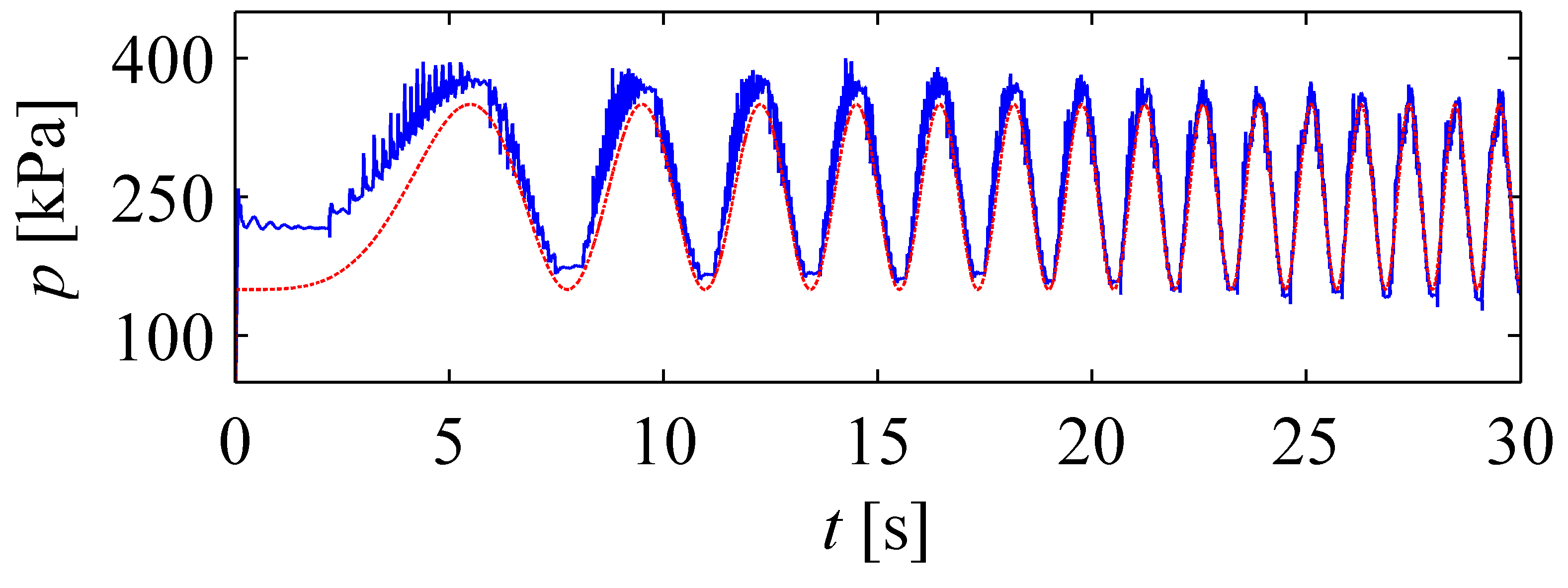}
\caption{Open-loop pressure control response: a measurement using a linear model for the soft robotic actuator and modified valve parameters.}
\label{fig:mbc_meas2}
\end{figure}

The feasibility of the model-based controller was also tested experimentally. The pressure inside the tube is controlled in open-loop without utilizing any measurements. The same sinusoidal reference signal is used as it was used in the simulation study. Fig~\ref{fig:mbc_meas1} shows that nonlinarities of the soft robotic actuator cause high error in pressure control accuracy. This experiment was carried out using presumable flow coefficients for the control valves calculated according to the orifice areas. However, Fig~\ref{fig:mbc_meas2} shows that, the pressure control performance significantly improves when valve parameters are modified in the controller. In this case, the flow coefficient of the high-pressure valve is increased for about $8\%$, whereas it is $58\%$ for the low-pressure valve.

\subsection{Closed-Loop Position Control}

In this part, the position is controlled using a position feedback provided from the vision system.

\subsubsection{Unloaded Scenario}

The objective of this part is to control the position of the robot tip while it is moving freely in X-Y plane in Fig.~\ref{fig:ExpSetup}. It should also be noted that X and Y coordinates of the tip are not independent of each other in this one segment and one-directional soft manipulator and thus, they can not be controlled independently. As a matter of fact, we refer to the Y coordinate of the manipulator's tip as tip position in this research.

\begin{figure}[t]
\centering
\includegraphics[trim={0 0 0 0},clip, width=8.0 cm]{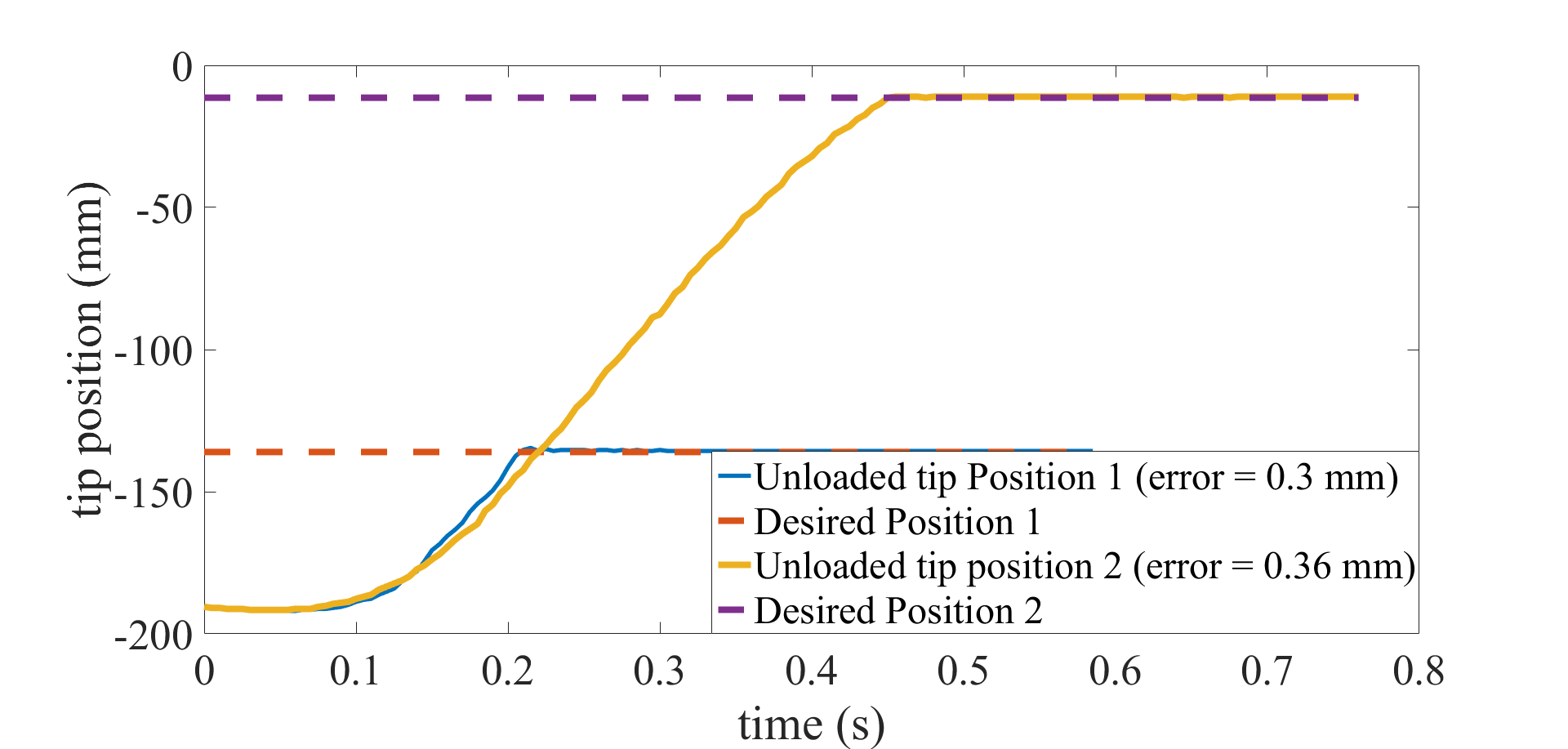}
\caption{Tip Position for different desired positions while unloaded}
\label{fig:TipPositionPositionControlUnloadedPressurize}
\end{figure}

\begin{figure}[t]
\centering
\includegraphics[trim={0 0 0 0},clip, width=8.0 cm]{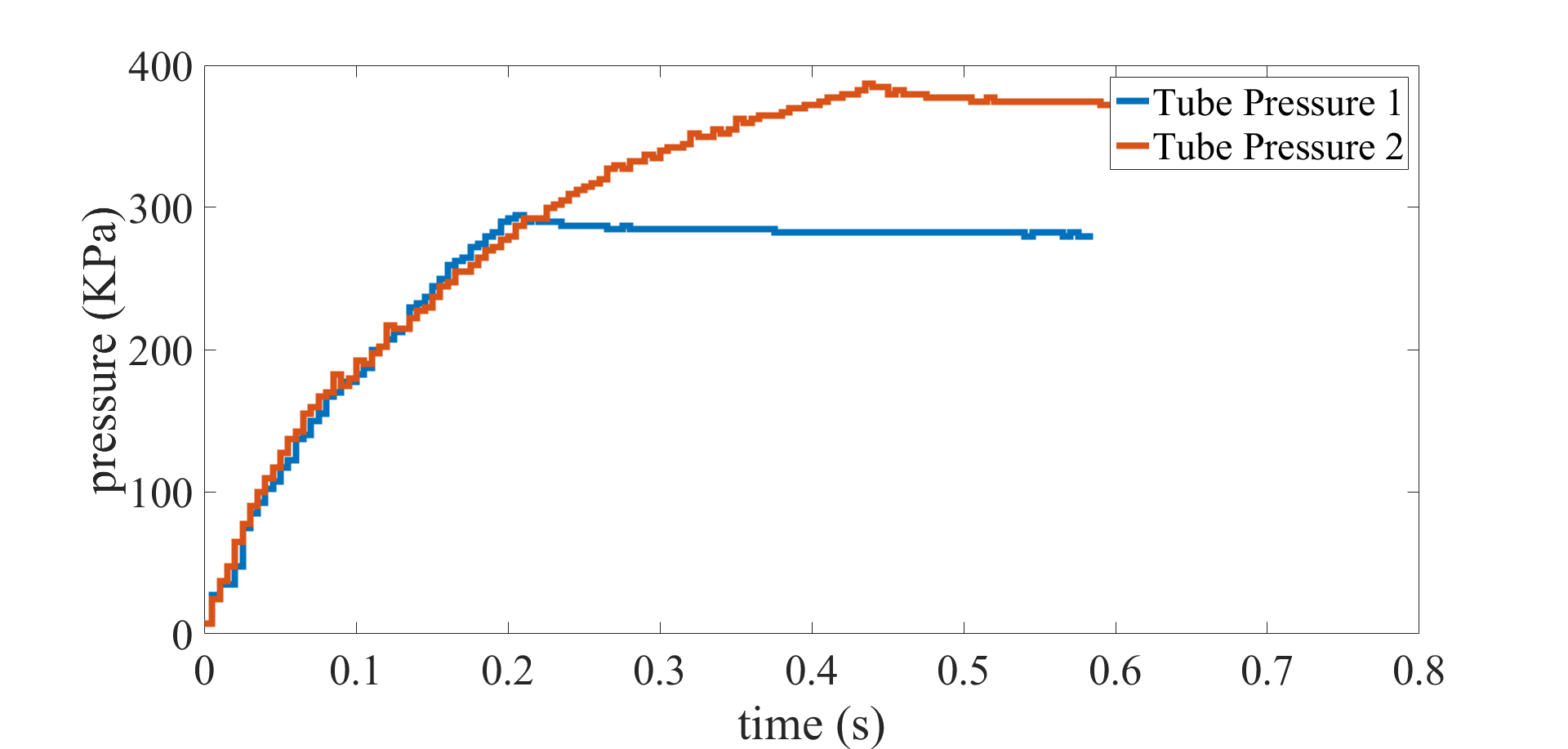}
\caption{Corresponding tube pressure for desired positions $1$, $2$, and $3$ in Fig.~\ref{fig:TipPositionPositionControlUnloadedPressurize}}
\label{fig:TubePressurePositionControlUnloadedPressurize}
\end{figure}

In order to verify the accuracy of the position control system, the manipulator was commanded to move to two different points in its workspace. For all two commands, the initial position is set to the point where the manipulator is fully depressurized. Fig. \ref{fig:TipPositionPositionControlUnloadedPressurize} shows the tip position while Fig.~\ref{fig:TubePressurePositionControlUnloadedPressurize} plots the corresponding tube pressure. Table \ref{tab:PositionControlUnloadedPressurize} reports the details of the experiments.

\begin{table}[h]
\caption{Manipulator data while unloaded}
\label{tab:PositionControlUnloadedPressurize}
\begin{center}
\begin{tabular}{c|c}
Points			& 	Position Error (mm)		\\
\hline
Desired Point 1	&	$0.3$				\\
Desired Point 2	& 	$0.36$			\\
\\\end{tabular}
\end{center}
\end{table}


\begin{figure}[t]
\centering
\includegraphics[trim={0 0 0 0},clip, width=8.0 cm]{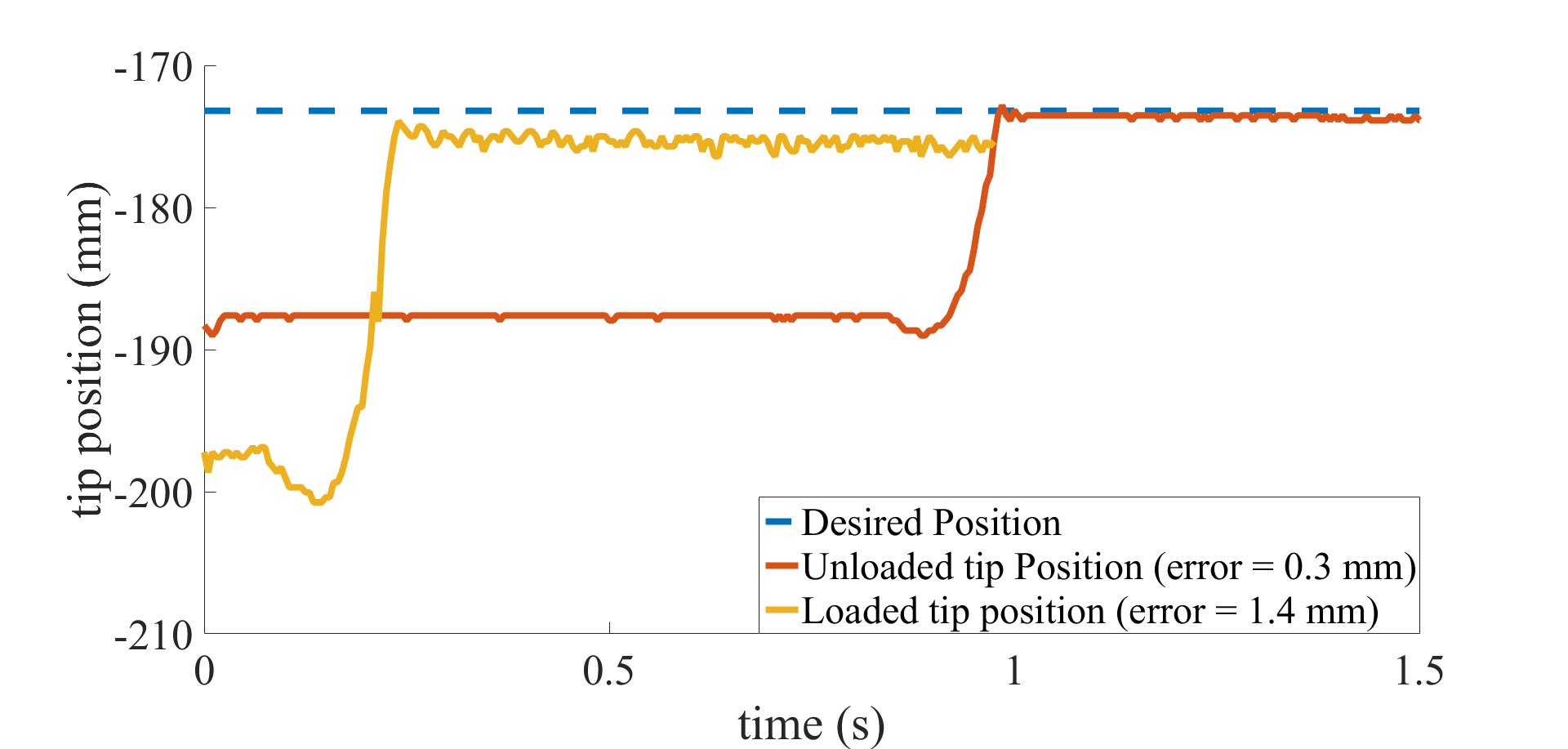}
\caption{Tip position while loaded and unloaded}
\label{fig:TipPositionPositionControlLoadedPressurize}
\end{figure}

\subsubsection{Loaded Scenario}

To verify the accuracy of the control system and also the soft robots performance while under external forces, a $10~g$ payload was attached to the tip of the robot. As it can be seen in Fig.~\ref{fig:TipPositionPositionControlLoadedPressurize}, the payload will change the initial condition. It should also be noted that for a given payload, there is a limit for the tip's position to where the manipulator can move the payload. Over the limit, the manipulator can not cope with the gravity/external forces and starts twisting around itself. To cope with this issue, some tube reinforcement techniques shall be considered or the design has to be modified.


\subsection{Position-Pressure Relationship}

\begin{figure}[t]
\centering
\includegraphics[trim={0 0 0 0},clip, width=8.0 cm]{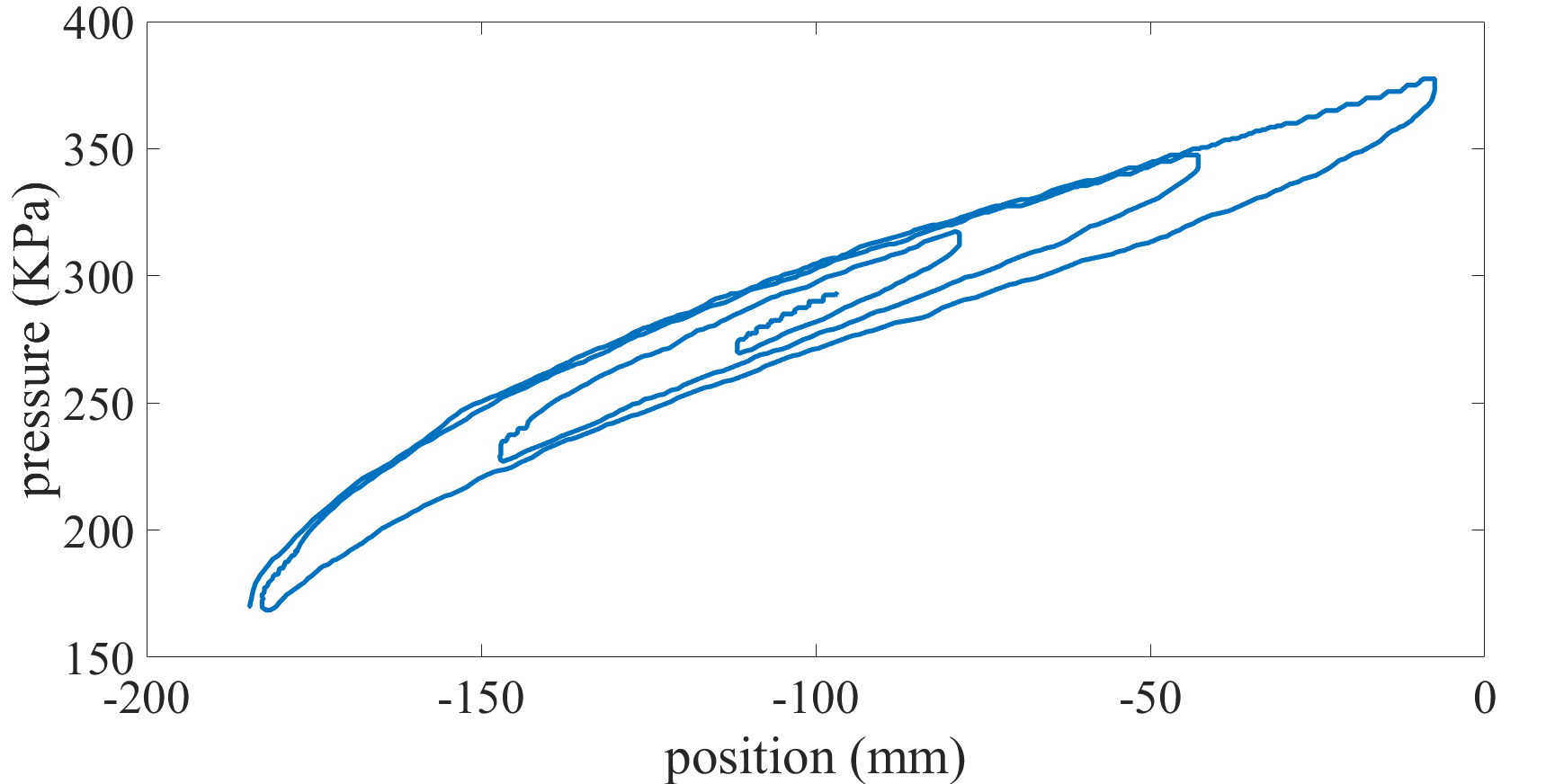}
\caption{System Hysteresis}
\label{fig:Hysteresis}
\end{figure}

Here we study if it is possible to establish a relationship between pressure and position of the tip. To do so, the manipulator is commanded to move to several predefined positions while the tip position and corresponding tube pressure is measured. Fig.~\ref{fig:Hysteresis} shows a hysteresis behavior of the system. It clearly shows that it is not possible to accurately open-loop control the tip position unless the initial conditions are known. However, the position-pressure relation, even not accurate enough has some advanteges. Due to heavy processing time, it may not be possible to track the tip position with camera as fast as the pressure can be measured. Yet, fusing them using the pressure-position model and Bayes or Kalman filter will provide us a faster position measurement rate.

\section{DISCUSSIONS AND CONCLUSIONS}

The developed model-based pressure control has quite fast response but moderate accuracy. On the other hand, the vision system has high accuracy but slow response because of the required processing time. Fusing the two approaches will result in a better accuracy and faster response. More importantly, the inner loop could be closed without the need for any pressure measurement sensor. This will be studied in our future work.

The experiments show that fast pressure control is feasible without the need for the pressure sensor using digital valves when the model-based approach is used. However, faster valves are needed to increase the control resolution. Furthermore, it has been shown that it is possible to accurately control the tip position with only an on/off controller for both loaded and unloaded scenarios. In addition to that, the hystersis of the system shows that it is not possible to control the tip position without any position feedback. Any position-pressure model should take into account the initial conditions or otherwise, the controller will fail to control the position. 

The future works of this research include working on improving the design to increase attained forces as well as enlarge controlled work space. We will also study these actuators in parallel and series configuration including their kinematic and dynamic modeling. Furthermore, developing and manufacturing fast miniature on/off valves is another step in the project. Besides, we aim towards more compact digital hydraulic drive system including control electronics.

\section*{APPENDIX}

To study the effect of different elastomer types, four structurally the same actuators were prepared from four different elastomers: $1)$ Polyurethane (PUR, ToppFlex from Oy Toppi Ab), $2)$ Polyolefine based thermoplastic vulcanizates (TPE, LKA $05/08$ from Teknikum Oy), $3)$ natural rubber (NR) and $4)$ polydimethylosiloxane (PDMS). The materials were selected to represent different hardnesses and thus different moduli. The Shore A hardness of the materials is presented in Table \ref{tab:ShoreA}. The hardness was measured according to ASTM D $2240-00$ with Affri Hardness tester. The NR was mixed in-house while other tubes were commercially available. The NR contained 45 phr N-234 carbon black as well as other ingredients such as curatives (sulphur, N-Cyclohexyl-2-Benzothiazolesulfenamide (CBS), zinc oxide, stearic acid), antioxidants (N-(1,3-dimethylbutyl)-N'-phenyl-1,4-Benzenediamine, 2,2,4-trimethyl-1,2-dihydroquinoline) and plasticizer (treated distillate aromatic extract oil).  The NR and its ingredients were mixed in a Krupp Elastomertechnik GK 1.5 E intermeshing mixer.  The CBS and sulphur were added on an open two-roll mixing mill and the mixing was continued until an even rubber belt was achieved. Then the tube (length $170~mm$, outer radius $8~mm$, inner radius $5~mm$) was formed by wrapping one $1.5~mm$ thick layer of rubber sheet around the metal rod (diameter $5~mm$). Flowing of the rubber during curing was prevented by wrapping a thin silicone sheet tightly around the tube. The other end was closed by a $15~mm$ long cap prepared from the same NR rubber compound. The tube was cured in a heating press at $\SI{150}{\degreeCelsius}$ for $20$ min.

\begin{table}[h]
\caption{Hardness of Eelastomers}
\label{tab:ShoreA}
\begin{center}
\begin{tabular}{c|c}
Elastomer		& 	Shore A \\
\hline
PUR			&	85  \\
TPE			& 	55  \\
NR			& 	45  \\
PDMS			& 	37\\\end{tabular}
\end{center}
\end{table}

\begin{figure}
    \centering
        \includegraphics[trim={0 0 0 0},clip, width=8.0 cm]{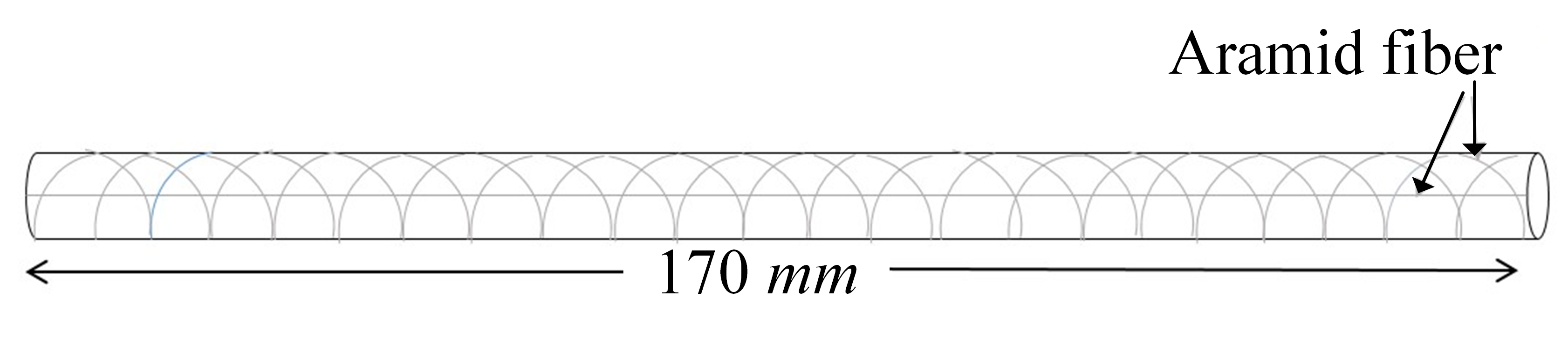}
        \caption{Design of the fiber-reinforced actuators}
        \label{fig:TubeSchematic}
\end{figure}

The ends of the commercial tubes (length $170~mm$, outer radius $8~mm$, inner radius $5~mm$) were closed by a $1.5~mm$ long cap fabricated from acrylic foam tape (VHB4910 from 3M). Next, an aramid fiber (Technora T-240 from Teijin Limited) was placed one side of the tube to to prevent longitudinal extension of tube during pressure applying. The radial expansion of the tube was restricted by hand wounding the same aramid fiber around the tube as presented in Fig.~\ref{fig:TubeSchematic}. The internal and external diameters of the tubes were chosen according to the potential application, in this case a wearable robot. Fig.~\ref{fig:ActuationPerformance-Turns} shows the force output and bending of PDMS with different number of fiber turns. For control part of the paper, number of turns of fibers was set to 240 as it gave the highest force output as well as bending angle.

The force output of the tubes was measured with a 10 N load cell (LTS-1KA, Kyowa). The load cell was installed above the tube and the changes in force during actuation were controlled via LabVIEW 2012. The pressure from 0.5 to 5$~KPa$ was applied into the actuator in 0.5$~KPa$ steps. The bending angle of the actuator at different pressures was determined from the transition of the end of the actuator. The results of the actuation measurements, i.e., the output force corresponding to a given pressure and and bending angle are presented in Fig.~\ref{fig:Force-Pressure} and Fig.~\ref{fig:Bending-Elastomer}, respectively. It can be seen that the TPE gives the highest actuation force at 5$~KPa$ while PUR has the poorest force output. PUR is very stiff material and would require higher pressure for higher force output by contrast to PDMS that yields to higher force output already at low pressures although the force output stays at constant above 2$~KPa$. Moreover, due to lower modulus of PDMS it bends more than the other elastomers. At 5$~KPa$, TPE withstands the highest force output showing that materials having higher modulus are able to produce higher forces than the materials with lower modulus. Furthermore, the bending angle of TPE and NR are still about $\SI{90}{\degree}$, thus in overall the best actuation performance is achieved with the elastomers having hardness about 50 Shore A.

\bibliographystyle{ieeetr} 
\bibliography{ICRA17_Bib}

\end{document}